\documentclass[fleqn,11p,times]{elsarticle}
\usepackage[utf8]{inputenc}

\usepackage{setspace}
\usepackage{graphicx}
\usepackage{graphics}
\usepackage{xcolor}
\definecolor{navy}{rgb}{0.000000,0.000000,0.501961}
\graphicspath{{Images/}}
\usepackage[labelfont=bf]{caption}
\usepackage{subcaption}

\usepackage[ruled,vlined]{algorithm2e}

\usepackage{amsmath}

\usepackage{amsthm}
\makeatletter
\def\els@aparagraph[#1]#2{\elsparagraph[#1]{#2\@addpunct{.}}}
\def\els@bparagraph#1{\elsparagraph*{#1\@addpunct{.}}}
\SetAlgoCaptionSeparator{.}
\RestyleAlgo{algoruled}

\makeatother

\usepackage{float}



\begin{document}

\begin{frontmatter}

\title{Solving a Stackelberg game on transportation networks in a dynamic crime scenario: a mixed approach on multi-layer networks}

\vspace{0.4cm}
\author[1]{Sukanya Samanta\corref{cor1}}
\ead{susamanta1@gmail.com}
\cortext[cor1]{Corresponding author}

\author[1]{Kei Kimura}

\author[1]{Makoto Yokoo}

\author[2]{Palash Dey}

\vspace{0.4cm}

\address[1]{Department of Informatics, Information Science and Electrical Engineering (ISEE), Kyushu University, Fukuoka, 819-0395, Japan}

\address[2]{Department of Computer Science and Engineering, Indian Institute of Technology Kharagpur, Kharagpur, West Bengal 721302, India}

\vspace{0.4cm}

\begin{abstract}
Interdicting a criminal with limited police resources is a challenging task in a dynamic crime scenario, as the criminal changes location over time. The vastness of the transportation network adds to the difficulty. To address this, we introduce the concept of a layered graph, where at each time step, a duplicate of the transportation network is generated to trace the possible movements of both the criminal/attacker and the police/defenders. We model this as a Stackelberg game, where the attacker seeks to escape from the network using one of the predefined exit points, while the defenders attempt to intercept the attacker on his escape route. Attacker strategy is generated by applying Dijkstra's algorithm on the layered networks. The attacker seeks to minimize, while the defenders aim to maximize the probability of interdiction. We propose an approximation algorithm on the layered networks to develop strategy for defenders. The efficacy of the developed approach is compared with the adopted MILP approach on large-scale transportation networks. We compare the results in terms of computational time and optimality gap. The quality of the results underscores the necessity of the developed approach, as it efficiently solves this complex problem in a short time frame.

\end{abstract}
\begin{keyword}
\texttt{Stackelberg game, Multi-layer time expanded network, Exact algorithm, Approximation algorithm, Resource allocation}
\end{keyword}
\end{frontmatter}

\section{Introduction}
\label{S:1}
We consider a Stackelberg game in which the defender is the leader and the attacker is the follower. The assumption is that the defenders have information only about the crime location in a large transportation network. The defenders attempt to capture the attacker before he can flee the city. We develop a novel mixed approach based on a layered graph concept to solve the Stackelberg model. This mixed approach employs an exact method to generate the attacker’s strategy and an approximation method to determine the defenders’ strategies. For the defender, the developed approximation algorithm finds the efficient movement of the defender from a random initial location. For the attacker, the initial strategy is any random path from the crime location to any exit point. This exact algorithm computes the optimal path for the attacker by applying Dijkstra’s algorithm on the network expanded over time.

The defender’s mixed strategy is updated based on the recalculated utility obtained from the restricted master problem (RestrictedStackelbergLP). The defender then commits to the resulting optimal mixed strategy, which is provided to the attacker. Given this committed strategy, the attacker computes a best-response path using the exact optimization approach. If the attacker generates a new strategy not previously included in the restricted set, it is added, and the process restarts from the master problem. Otherwise, the defender attempts to generate a new pure strategy using the developed greedy algorithm on the time-expanded network. If a new defender strategy is found, it is added to the restricted set, and the procedure restarts. The algorithm continues iteratively, checking whether either player can produce a new improving strategy. If neither the attacker nor the defender can generate a new best response, the convergence criterion is satisfied and the algorithm terminates. The defender’s equilibrium utility is then evaluated based on the final best-response strategies of both players. The final defender utility at equilibrium and the total computational time are reported. For benchmarking purposes, the proposed Stackelberg framework is also solved using the adopted MILP formulations for both the attacker and the defender to evaluate the optimality gap.

We consider a Stackelberg game instead of a zero-sum game. The advantage of the Stackelberg game over the zero-sum game is that it allows for a more realistic representation of real-world scenarios where one player has more information, resources, or power than the other player. In this Stackelberg game, the defenders have an advantage because they can consider the attacker's potential moves and responses when making their decisions. This can result in a more efficient outcome than in a zero-sum game, where the players are equally matched.

Due to the complex transportation network and limited police resources, generating efficient strategies for both players is a challenging task. This paper introduces a Stackelberg game model for dynamic crime scenarios, presenting a novel exact algorithm for the attacker and a novel approximation algorithm for the defenders on a multi-layer network. The proposed MLN-EIGS approach produces high-quality solutions comparable to the adopted MILP-based approach while significantly reducing computational time.

The key novelty of this work is the integration of dynamic probabilistic interdiction modeling with a Stackelberg security game framework on a time-expanded transportation network. While prior works have studied static graph-based security games or MILP-based interdiction models, they do not simultaneously capture (i) temporal feasibility of movements, (ii) multiplicative interception risk along escape paths, and (iii) scalable equilibrium computation on large real-world networks. By combining a layered network representation with a logarithmic reformulation of escape probabilities and a restricted strategy Stackelberg solution framework, the proposed approach enables exact follower optimization via shortest-path methods while preserving the leader–follower structure. This modeling–algorithmic integration allows the framework to scale to large transportation networks where classical MILP formulations become computationally prohibitive.

The paper is organized as follows. In Section 2, we present the relevant research. In Section 3, we define the problem description and modeling. In Section 4, we describe the solution methodology. The benchmarking algorithm is described in Section 5. Section 6 presents the quality of the results. We conclude this research in Section 7.

\section{Related work}
\label{S:2}

Security games play an important role in providing social security (\cite{hunt2024review}, \cite{samanta2022literature}). Recent literature focuses on Stackelberg games, considering the escape interdiction problem to decrease the crime rate in society. For example, \cite{zychowski2023coevolution} consider search games (SEG) on directed graphs. They consider multiple defender resources and one attacker, where the attacker aims to reach one of several predefined target points from a fixed location. They develop a genetic algorithm-based heuristic approach to provide a near-optimal solution on synthetic datasets. \cite{basilico2009leader} consider a leader-follower game and formulate the problem as a mathematical programming model. They use optimization software tools to solve the model by generating the optimal strategies for both leader and follower. Again, \cite{letchford2013solving} consider security games on graphs and develop a polynomial-time algorithm to generate optimal strategies for players. Similarly, \cite{iwashita2016simplifying} consider security games on graphs and develop an algorithm to reduce the graph by eliminating unnecessary edges, providing a time-efficient, scalable near-optimal solution. In the same vein, \cite{karwowski2019monte} consider an evader-defender Stackelberg game model and develop a Monte Carlo Tree Search approach to provide efficient patrolling schemes. In addition, \cite{wang2019repeated} introduce a repeated Stackelberg security game that incorporates a cooperative human behavior model to enhance patrolling strategies for wildlife protection by modeling human decision-making in repeated interactions, thereby improving defender effectiveness against adaptive adversaries in dynamic environments.

\cite{tsai2010urban} consider an attacker-defender Stackelberg game model and develop a linear program to generate optimal mixed strategies for players by allocating limited resources optimally, with a case study on part of the Mumbai road network. Likewise, \cite{shieh2012protect} develop a game-theoretic system to provide security with limited resources in the port of Boston and also test its efficacy in the port of New York. They schedule patrols efficiently to provide optimal mixed strategies for players, considering an attacker-defender Stackelberg game model. Similar papers focus on Stackelberg security games to provide security in society (e.g., \cite{cermak2016using}, \cite{lou2017multidefender}, \cite{sinha2018stackelberg}, \cite{vcerny2018incremental}, \cite{zhang2021bayesian}).

\cite{conitzer2006computing} show that exact approaches are not suitable due to the NP-hard nature of this problem, as it faces scalability issues due to the high time complexity. On the contrary, \cite{paruchuri2008playing} develop a MILP-based exact algorithm to solve Bayesian Stackelberg security games. Similarly, \cite{bosansky2015sequence} consider Stackelberg games and develop algorithms based on LPs and MILP to generate Strong Stackelberg Equilibrium (SSE) and perform a case study at Los Angeles International Airport, focusing on security scheduling.

Considering a zero-sum game for the escape interdiction problem, \cite{zhang2017optimal} develop a MILP-based solution approach to provide an optimal solution. To demonstrate the efficacy of their methodology, they generate optimal solutions on grids of different sizes. \cite{samanta2022vns} consider the same zero-sum game problem and develop a meta-heuristic-based solution approach to provide a scalable near-optimal solution in a time-efficient manner. Again, \cite{samanta2021vehicle} develop a simulation-based approach to generate a scalable solution to increase security in a large transportation network for this escape interdiction problem.

The layered graph concept is a useful tool for solving complex problems on transportation networks in a time-efficient manner. For example, \cite{saito2009discovering} consider the problem of selecting important nodes in a network and construct a layered graph from the original graph, where each layer is added on top as time proceeds to demystify the complex problem. We focus on solving a Stackelberg escape interdiction game on large-scale transportation networks using a layered graph concept.

\section{Problem description and modeling}
\label{S:3}

We consider a two-player Stackelberg game, where the sequential interaction occurs between multiple defenders $\overline{D} = \{d_r \mid r \in R\}$ and a single attacker $\overline{A}$. The total number of defenders is $m$, and the set of all defenders is represented by $\overline{D}$. Here, $r \in R = \{1, \dots, m\}$. Since the defenders act jointly and a complete defender strategy comprises the strategies of all individual defenders, while there is only one attacker, we refer to this as a two-player Stackelberg game. 

The defender has a finite set of pure strategies $\Acute{S}$, and the attacker has a finite set of pure strategies $\Acute{A}$. Let $x$ and $y$ denote the corresponding mixed strategies, i.e., probability distributions over $\Acute{S}$ and $\Acute{A}$, respectively.

The transportation network is represented as a directed graph $G = (V, E)$, where $E$ is the set of directed edges corresponding to roads, and $V$ is the set of nodes representing intersections. There is a set of predefined exit points in the considered network. $v_\infty$ signifies any exit node in the considered network. The game begins at time 0 and ends at time $t_{\text{max}} > 0$. 

The sequence of states $A = \langle a_1 = (v^a_0, 0), \ldots, a_j = (v_j, t^a_j), \ldots, a_k = (v_\infty, t^a_k \leq t_{\text{max}}) \rangle$ represents the pure strategy of the attacker. Each state $a_j = (v_j , t^a_j )$ indicates that at time $t^a_j$, the attacker is present at node $v_j$. Likewise, a defender's state $d_r$ is a tuple $S^r = (v^r, t^{r,in}, t^{r,out})$, representing the state where defender $d_r$ is present at the node $v^r$ during the interval $[t^{r,in}, t^{r,out}]$.

The defender's pure strategy is denoted by $S$. A pure strategy for the defender consists of $m$ schedules, i.e., $S = \{S^r : r \in R\}$. The schedule for defender $d_r$ is defined as a sequence of states $S^r = <s^r_1 , . . . , s^r_i , . . . , s^r_k >$, where $s^r_1 = (v^r_0 , 0 , t^{r,out}_1)$ and $s^r_k= (v^r_k , t^{r,in}_k , t_{max})$. The mixed strategy for the defender is denoted by $x = <x_S>$, where $x_S$ represents the probability with which the strategy $S$ is played.

For $a_j = (v_j, t^a_j )$ and $s^r_i= (v^r_i , t^{r,in}_i , t^{r,out}_i)$, the defender $d_r$ intercepts the attacker at node $v^r_i$ if $v^r_i$ = $v_j$ and $t^{r,in}_i \leq t^a_j \leq t^{r,out}_i$. Since the defender receives a payoff of 1 upon interception and 0 otherwise, the expected utility under mixed strategies is equal to the probability that the attacker is intercepted. Correspondingly, the attacker’s utility is defined as the probability of successfully escaping the network.

The defender’s optimal strategy is obtained by solving the linear program given in Eqs. (1)–(3).

\begin{equation}
max \qquad U
\end{equation}
\begin{equation}
s.t. \qquad U \leq U_d \; ( x, A ) \qquad \forall A \in \Acute{A}
\end{equation}
\begin{equation}
\sum\limits_{ S \in \Acute{S} } x_s = 1, x_s \geq 0 \qquad \forall S \in \Acute{S}
\end{equation}


In Eq. (4), the defender commits to a strategy $x \in \Acute{S}$, and given such an $x$, the attacker chooses his strategy from the best-response set $BR(x)$ where

\begin{equation}
BR (x) = \; argmin_{y \in \Acute{A}} \; U_d \; ( x, y ) 
\end{equation}

To maximize utility, the defender chooses the strategy $x$ (the best response of the defender) to the attacker's best response (see Eq. 5). 

\begin{equation}
max_{x \in \Acute{S}} \; U_d \; ( x, y ) \qquad	s. t. \; y \in BR (x)
\end{equation}

Strong Stackelberg Equilibrium (SSE) is popular because it is always guaranteed to exist (\cite{kroer2022lecture}). In case of a SSE, we assume that the follower (attacker) breaks ties in favor of the leader (defender) (see Eq. 6). In that case, the optimization problem is
\begin{equation}
max_{x \in \Acute{S}, \; y \in BR (x)} \; U_d \; ( x, y ) 
\end{equation}

\section{Proposed mixed approach on multi-layer networks}
\label{S:4}
This section presents the proposed solution methodology for the considered escape interdiction problem which is formulated as a Stackelberg game model. A novel mixed approach on layered graph concept is developed to solve the Stackelberg model, where the defender is the leader and the attacker is the follower. In this Stackelberg game, first, the defenders commit to a strategy, and then the attacker generates the best response to the given defender strategy. We use the concept of a multi-layer network (MLN) in which, at each time-stamp, we create a copy of the entire network.
We consider edge length as the time factor to create connections between these multi-layer networks. This means that depending on the edge length, we choose the layers from which the start and end nodes of that particular edge are selected. In this way, for all edges in the original transportation network, we create corresponding edges in the multi-layer network.
We consider a set of strategies for both players, i.e., the defenders and the attacker. For the defender, we use a mixed strategy set where each strategy is assigned a mixed probability, with the sum of these probabilities equaling one. For the attacker, we consider a pure strategy set.

To compute the attacker’s optimal strategy, node weights are determined based on the defender’s mixed strategy. The attacker is considered intercepted if interdiction occurs at any node along the selected path; that is, interception corresponds to the union of interdiction events across all visited nodes. The resulting interdiction probability $P(n,t)$ is assigned to each node and incorporated into the weights of its incoming edges. Using the logarithmic transformation of the escape probability, Dijkstra’s algorithm is then applied to identify the attacker’s path that minimizes the overall probability of interdiction.

The MLN representation is essential for accurately modeling the temporal dimension of the escape interdiction problem. Since edge lengths represent travel time, both attacker and defender strategies are inherently time-dependent, and feasibility depends not only on spatial connectivity but also on the timing of node visits. A single-layer graph cannot distinguish between paths that traverse the same node at different times, leading to incorrect evaluation of interdiction risk and strategy feasibility. The MLN explicitly encodes time by creating copies of the network at discrete time steps and connecting them according to edge travel times, thereby transforming the time-dependent path-planning problem into a static shortest-path problem. This representation enables the exact computation of the attacker’s best response using Dijkstra’s algorithm and allows the defender to correctly anticipate temporally feasible attacker strategies. Moreover, the MLN facilitates the aggregation of interdiction probabilities under mixed defender strategies in a principled manner, ensuring both modeling accuracy and computational tractability.

RestrictedStackelbergLP computes the best mixed strategy of the defender $x^{\star}$, by solving the linear program defined in Eqs. (1)–(3), using the complete strategy spaces of the defender and attacker, denoted by $\Acute{S}$ and $\Acute{A}$, respectively, as input. Here, $ExactAO$ denotes the exact approach developed for the attacker, while $ApproxDO$ denotes the approximation algorithm developed for the defender.

For the defender, we create a multi-layer network in the same way as for the attacker. Weights are assigned to all nodes based on the attacker strategies, with each attacker strategy given equal probability. We develop a novel approximation algorithm for the defender to generate a near-optimal defender strategy. Thus, for the defender, the problem can be considered as finding a near-optimal path with the aim of covering at least one node from each attacker strategy. For this, we use different colors to represent each attacker strategy. The developed defender strategy includes as many different colored vertices as possible.

The algorithm terminates when neither player can generate an improving strategy outside the restricted sets, ensuring convergence to a approximate Stackelberg equilibrium of the full game (see Algorithm~\ref{alg:a1}).

\begin{algorithm}[H]
\SetAlgoLined
 \textbf{Input:} Initialize the initial strategy sets of defender and attacker $S', A'$\;
 \textbf{Output:} $U^{\star}_d:$ Defender's utility\;
 \Repeat{convergence (no new strategies)} {
 
        $(x^{\star}) \leftarrow RestrictedStackelbergLP (S^{'}, A^{'})$\;

        Defender commits to the best mixed strategy $x^{\star}$\;
            
		$ BR: \; A^{\star} \leftarrow { ExactAO(x^{\star}) }$\;

		\If{$A^{\star} \notin A^{'} $}{
   			$A^{'} \leftarrow A^{'} \cup \{A^{\star}\}  $\;
            \textbf{continue}\;
   			}

 		$ S^{\star} \leftarrow {ApproxDO(A^{'}) }$\;

		\If{$S^{\star} \notin S^{'} $}{
   			$S^{'} \leftarrow S^{'} \cup \{S^{\star}\}  $\;
            \textbf{continue}\;
   			} 
    }
$U_d^{\star} \leftarrow U_d(x^{\star}, A^{\star})$\;
\tcp{Final defender utility at equilibrium}

\Return $U_d^{\star}, (x^{\star}, A^{\star})$.

 \caption{Stackelberg game solved using the MLN-EIGS algorithm.}
 \label{alg:a1}
\end{algorithm}

The proposed model constitutes a Stackelberg (leader–follower) game rather than a simultaneous-move game because the defender commits to a mixed strategy before the attacker selects its path. Given this commitment, the attacker computes an exact best response using the multi-layer network representation. The defender’s optimization explicitly anticipates this best response, resulting in a bilevel structure of the form $max_{x} \; U_d\big(x, BR(x)\big)$. This sequential decision process, together with the follower’s best-response computation after observing the leader’s commitment, characterizes a Stackelberg equilibrium obtained via backward induction. In contrast, a simultaneous (Nash) formulation would require both players to select strategies without prior observation and would not involve a nested best-response structure. The problem is formulated as a bilevel optimization model, which is equivalent to a Stackelberg game.

The Stackelberg approximate equilibrium is achieved through an iterative restricted strategy expansion framework that preserves the leader–follower hierarchy. At each iteration, the restricted Stackelberg game defined over the current defender and attacker strategy sets $S^{'}$ and $A^{'}$ is solved using RestrictedStackelbergLP to obtain the defender’s optimal mixed strategy $x^{\star}$. The defender commits to this strategy, after which the attacker computes a best response over its full strategy space via $ExactAO(x^{\star})$. If this best response is not already included in $A^{'}$, it is added to the attacker’s strategy set and the restricted game is re-solved. Otherwise, the defender generates an improving strategy over its full strategy space using $ApproxDO(A^{'})$; if a new strategy is found, it is added to $S^{'}$ and the process repeats. The algorithm terminates when neither player can generate a profitable deviation outside the restricted sets. At this point, no attacker strategy yields a higher payoff against the committed defender strategy, and no defender strategy improves its utility anticipating the attacker’s best response.

Since the defender’s strategy generation relies on an approximation algorithm rather than an exact optimization procedure, the leader’s optimality condition may not be satisfied globally. Consequently, the proposed method computes an approximate Stackelberg equilibrium, where the attacker plays an exact best response but the defender’s strategy is only approximately optimal. Therefore, the resulting pair $(x^{\star}, A^{\star})$ satisfies the approximate Stackelberg equilibrium conditions, and $U_d^{\star}$ represents the defender’s approximate Stackelberg equilibrium utility.

\subsection{Efficient attacker strategy design using exact approach on time expanded network}
\label{S:4.1}

The optimal attacker strategy is computed using an exact approach (see Algorithm~\ref{alg:a3}). For each time-stamp, a separate copy of the entire network is created, forming a multi-layer structure. Connections between these layers are established based on the edge lengths in the original graph. Initially, all edge weights in the multi-layer network are set to zero. The algorithm outlines the process of assigning interdiction probabilities to each node and the corresponding weights to its adjacent edges. Finally, Dijkstra’s algorithm is applied to this multi-layer network to determine the attacker’s optimal strategy.

In computing the attacker’s best response, it is essential to model the probabilistic structure of sequential interdiction correctly. The attacker is intercepted if interdiction occurs at \emph{any} node along the selected path, which corresponds to the union of interception events over all visited nodes. The interdiction probability at a node $(n,t)$ equals the total probability mass of defender strategies that occupy that node at time $t$. The attacker escapes only if interception does not occur at every visited node. Under the standard assumption that interdiction events across distinct time steps are independent, the escape probability along a path $A$ is therefore given by
\[
\prod_{(n,t)\in A} \left(1 - P(n,t)\right),
\]
and the corresponding interception probability is
\[
1 - \prod_{(n,t)\in A} \left(1 - P(n,t)\right).
\]
To maintain computational tractability while preserving this multiplicative probabilistic structure, we apply a logarithmic transformation and assign weights $w(n,t) = -\log\left(1 - P(n,t)\right)$ to the layered network. This transformation converts the product of escape probabilities into a summation of nonnegative edge weights, thereby enabling the use of Dijkstra’s algorithm to compute the path that truly minimizes the interception probability.

\begin{algorithm}[H]
\SetAlgoLined
 \textbf{Input:} Crime node is the “START” node, all exit nodes are the “GOAL” nodes, Original graph $(G_0)$\;
 \textbf{Output:} Best attacker strategy having minimum probability of interdiction\;
 \textbf{Construction of layered graphs:}\\
 \For{$( i = 0; i < t_{max}; i++)$}{
    Generate one copy of the original graph/network, labeled as $G_{i+1}$ \;
 }
\textbf{Connect the Layered Graphs depending on time/distance between adjacent nodes in the original graph:}\\
\For{all edges in the original graph}{
    $L$ = current edge length, $S$ = From node, $T$ = To node (in original graph) \;
    \For{$(j = 0; j < t_{max} - L; j++)$}{
    Select graph $G_{j}$ and graph $G_{j + L}$ \;
    Create edge from node $S$ of  $G_{j}$ to node $T$ of  $G_{j + L}$ \;
    }
    }
\textbf{Update the maximum probability of interdiction for each node:}\\

Initial probability of interdiction $(P)$ of all nodes in the layered graph = 0 \;
Initial weight of all edges in the layered graph = 0 \;
\caption{Optimal attacker strategy design using time expanded network.}
 \label{alg:a3}
\end{algorithm}

\begin{algorithm}[H]
\SetAlgoLined

\For{all $N$ defender strategies in the best strategy set}{
    \For{all nodes present in the current strategy}{
        $Node = n, t_{in} = In \;Time, t_{out} = Out\; Time$
            \While {$t_{in} < t_{out}$} {
            Select node $n$ of graph $G_{t_{in}}$\;
            Update the maximum probability of Interdiction of that node $(n, G_{t_{in}})$:\\
            $P$ = $w(n, G_{t_{in}}) =P + (-\log(1 - P_{mix}))$ where $P_{mix}$ is the mixed prob. of the current defender strategy.\;
            $t_{in} = t_{in} + 1$\;
            }       
        }
    }
\textbf{Assign weight to all incoming edges of a node:}\\
\For{all nodes present in the layered graph}{
    \If{Probability of interdiction of a node $(n, G_{t_{in}})$ is $w(n, G_{t_{in}})$}{
        Assign the weight $w(n, G_{t_{in}})$ to all in-coming edges of that node.
    }
}
\textbf{Apply Dijkstra’s algorithm on the developed Multi-Layer Network.}
 \Return The optimal attacker path with the minimum probability of interdiction.
\end{algorithm}

After assigning weights $w(n,t) = -\log\left(1 - P(n,t)\right)$ to each node in the layered network, Dijkstra’s algorithm is applied to compute the path with minimum total weight. Since minimizing 
\[
\sum_{(n,t)\in A} -\log\left(1 - P(n,t)\right)
\]
is equivalent to minimizing the interception probability 
\[
1 - \prod_{(n,t)\in A} \left(1 - P(n,t)\right),
\]
the resulting shortest path corresponds to the attacker strategy with minimum probability of interdiction.

\subsection{Efficient defender strategy design using approximation algorithm on time expanded network}
\label{S:4.2}

We develop a novel approximation algorithm to generate a near-optimal strategy for the defender in a time-efficient manner. Since the attacker's strategy set consists of pure strategies, we assign different colors to each of these attacker strategies. Then, we attempt to construct a path for the defender that includes at least one colored vertex to interdict each attacker strategy. This implies that the defender's strategy contains as many different colored vertices as possible. We describe the approach in the following steps.

\begin{itemize}
\item \textbf{Input} 
\begin{itemize} 
\item A directed acyclic weighted graph, where some vertices are colored.
\item There are $k$ different colors $(c_1, c_2, ..., c_k)$.
\item A threshold value of the path length $t$.
\end{itemize}
Here, each colored vertex corresponds to a particular attacker's strategy.
\item \textbf{Goal:} \\
To find a defender's strategy whose length is at most $t$ and contains as many different colored vertices as possible.

\item \textbf{Guess:} \\
This problem is difficult (say, its decision version is $NP$-complete) and thus cannot be solved by Dijkstra. We provide a formal proof in this paper.

\end{itemize}

We use the steps below for the approximate algorithm.

\begin{itemize}
\item \textbf{Step1:} Find a shortest path to \textit{any} one of colored vertices from the start vertex using Dijkstra. Assume the path is to vertex $v_1$ with color $c_{i_1}$.

\item \textbf{Step2:} Find a shortest path to \textit{any} one colored vertices except $c_{i_1}$ from $v_1$. Assume the path is to vertex $v_2$ with color $c_{i_2}$.

 \item \textbf{Step3:} Find a shortest path to \textit{any} one colored vertices except $c_{i_1}$ and $c_{i_2}$ from $v_2$, and so on, until all colors are visited or the total path length reaches $t$. 

\end{itemize}

In the simplest form, if the path we obtained in the previous method does not cover a subset of colors, we create another path from the initial vertex, which tries to cover these remaining colors only. We repeat this procedure until all colors are covered. Then, the defender flips a coin and chooses one path.

\subsection{Problem ‘color covering’ is $NP$-complete}
\label{S:4.3}

\textbf{Problem ‘color covering’}
\begin{itemize}
\item \textbf{Input:} A directed graph and the initial vertex. Some vertices are colored. There are $m$ different colors. It is possible that one vertex has multiple colors.
\item \textbf{Output:} ‘Yes’ if there exists a path from the initial vertex with length $n$, such that all $m$ colors appear on at least one vertex along the path. ‘No’ otherwise.
\end{itemize}

\textbf{Theorem:} Problem ‘color covering’ is NP-complete
\begin{itemize}
\item \textbf{Proof idea:} Reduction from 3-SAT (which is known to be NP-complete).
\end{itemize}

\textbf{Problem 3-SAT}
\begin{itemize}
\item \textbf{Input:} $n$ boolean variables $(x_1, \ldots, x_n)$, $m$ clauses. Each clause is a disjunction of three literals. Each literal is a variable or its negation.
\item \textbf{Output:} ‘Yes’ if there exists an assignment of variables that makes all clauses true. ‘No’ otherwise.
\end{itemize}

For a given $3-SAT$ instance, we create an $n+1$ level network.
\begin{itemize}
\item There is one level-$0$ vertex, which is the initial vertex.
\item There are two level-$i$ vertices (for $i>0$). One vertex corresponds to making variable $x_i$ true. The other vertex corresponds to making variable $x_i$ false.
\item There exists a directed edge from each of level-$i$ vertex to each of level-$(i+1)$ vertices.
\item Each clause has its own color.
\item If a clause with color $c$ contains $x_i$,  the ‘true’ vertex for $x_i$ has color $c$.
\item If a clause with color $c$ does not contain $ x_i$,  the ‘false’ vertex for $x_i$ has color $c$.
\end{itemize}
Thus, the $3-SAT$ instance is satisfiable iff there exists a path from the initial vertex with length $n$, which covers all colors.

\section{MILP-EIGS benchmarking algorithm}
\label{S:5}

To establish a benchmark, we formulate a Stackelberg game analogous to the MLN-EIGS framework. In this formulation, the optimal strategies for both the attacker and the defender are obtained using $bestAo$ and $bestDo$, which correspond to the MILP-based solution approaches for the attacker and the defender, respectively (see Algorithm~\ref{alg:a2}). These optimal approaches, developed by \cite{zhang2017optimal}, compute the best strategies for attackers and defenders, given a predefined strategy set for each player. 

The vehicle interdiction problem is proved to be NP-hard (\cite{zhang2017optimal}). The best oracles, that is, the MILPs, encounter significant space and time complexity when applied to moderately large urban road networks with many nodes and edges. The MILP consists of bestDo for defenders and bestAo for the attacker. \cite{zhang2017optimal} develop these MILP approaches to find the best strategies.

The MILP approach for the attacker (bestAo) constructs an optimal path for the attacker from the crime node to the exit node. The attacker’s utility decreases when more defender paths interdict this new attacker path. The MILP approach for defenders (bestDo) focuses on maximizing the rate at which the defender can intercept the attacker within a specified time frame in a large transportation network. The bestDo provides the optimal formulation for the defender’s movements over time. It devises a path for the defender that intercepts the maximum number of attacker paths, thereby maximizing the defender’s utility.

Since both the attacker’s best response and the defender’s strategy generation are solved exactly, the final strategy pair $(x^{\star}, A^{\star})$ satisfies the optimality conditions of the leader–follower model and therefore constitutes a Stackelberg equilibrium of the full game, with $U_d^{\star}$ representing the defender’s equilibrium utility. To evaluate the effectiveness of the proposed MLN-EIGS approach in terms of optimality gap and computational efficiency, we compare the defender’s utility and computation time with those obtained using the benchmark MILP-EIGS method.

\begin{algorithm}[H]
\SetAlgoLined
 \textbf{Input:} Initialize the initial strategy sets of defender and attacker $S', A'$\;
 \textbf{Output:} $U^{\star}_d:$ Defender's utility\;
 \Repeat{convergence (no new strategies)} {
 
        $(x^{\star}) \leftarrow RestrictedStackelbergLP (S^{'}, A^{'})$\;

        Defender commits to the best mixed strategy $x^{\star}$\;
            
		$ BR: \; A^{\star} \leftarrow { bestAo(x^{\star}) }$\;

		\If{$A^{\star} \notin A^{'} $}{
   			$A^{'} \leftarrow A^{'} \cup \{A^{\star}\}  $\;
            \textbf{continue}\;
   			}

 		$ S^{\star} \leftarrow {bestDo(A^{'}) }$\;

		\If{$S^{\star} \notin S^{'} $}{
   			$S^{'} \leftarrow S^{'} \cup \{S^{\star}\}  $\;
            \textbf{continue}\;
   			} 
    }
$U_d^{\star} \leftarrow U_d(x^{\star}, A^{\star})$\;
\tcp{Final defender utility at equilibrium}

\Return $U_d^{\star}, (x^{\star}, A^{\star})$.
 \caption{Stackelberg game solved using the MILP-EIGS benchmarking algorithm.}
 \label{alg:a2}
\end{algorithm}

\section{Results and discussion}
\label{S:6}
In this section, we present the results of the developed approaches. The proposed algorithms are coded in Python 3.6 and tested on a computer equipped with an Intel(R) Core(TM) 3.20 GHz processor and 8 GB RAM, operating under the LINUX environment. All MILPs are solved using CPLEX (version 12.8).

\begin{figure}[H]
\centering
\includegraphics[scale=0.35]{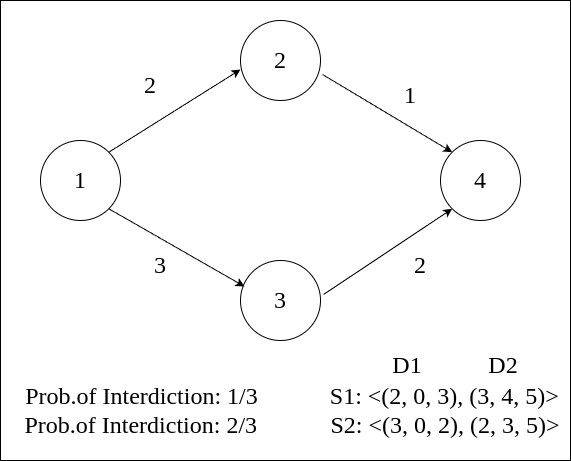}
\caption{Sample network for designing the optimal attacker strategy.}
\label{image-1}
\end{figure}

\begin{figure}[H]
\centering
\includegraphics[scale=0.36]{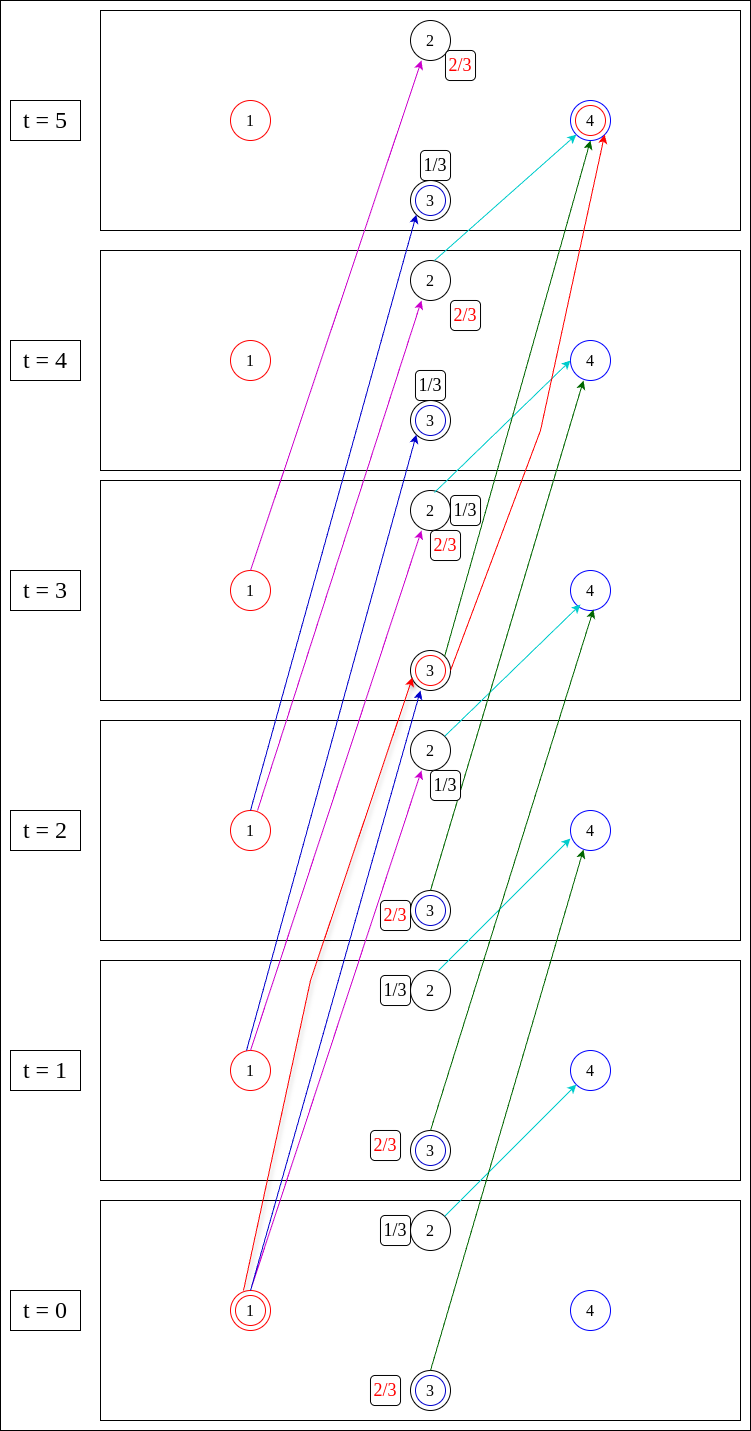}
\caption{Design of a multi-layer network for attacker considering a sample network (Fig. \ref{image-1}).}
\label{image-4}
\end{figure}

In Fig. \ref{image-1}, we consider a sample network of 4 nodes in which police stations are nodes 2 and 3, the crime node is 1, the maximum time limit ($t_{max}$) is 5, and the exit point is node 4. Here, $0\_4$ indicates node 4 at timestamp 0 $(t=0)$ in the multi-layer network. In this example, we provide two mixed defender strategies as input with probabilities of $1/3$ and $2/3$. Each node in this multi-layer network is assigned a corresponding probability of interdiction. To generate the optimal attacker strategy, we use Dijkstra's algorithm on the time-expanded network (see Fig. \ref{image-4}). The final attacker strategy is represented by the red line in the multi-layer network, which follows the path $0\_1 \rightarrow 3\_3  \rightarrow 5\_4$. We demonstrate that our developed exact approach for the attacker can generate the optimal attacker strategy, enabling the attacker to escape without interdiction in a concise amount of time (see Table \ref{tab:table1}).

\begin{table}[H]
  \begin{center}
   \caption{\\Optimal attacker strategy design using Dijkstra algorithm on time expanded network.}
    \label{tab:table1}
    \scalebox{0.8}{
    \begin{tabular}{llll} 
    \hline
      \multicolumn{4}{c}{\textbf{Game Parameters}} \\ \hline
      \multicolumn{4}{c}{\textbf{Network Size: 4 Nodes, Crime Node: $0\_1$, Police Stations: $0\_2$, $0\_3$, $T_{max}$: 5, Exit Point: 4}}\\ \hline
      \parbox[t]{2cm}{Test Case} &  \parbox[t]{4cm}{Optimal Attacker strategy} & \parbox[t]{4cm}{Utility of the Final\\  Optimal Attacker Strategy} & \parbox[t]{2cm}{Run Time \\(Sec)}\\ \hline
       1 & [$0\_1, 3\_3, 5\_4$] & 0.0 & 0.004\\ \hline
       2 & [$0\_1, 3\_3, 5\_4$] & 0.0 & 0.004\\ \hline
       3 & [$0\_1, 3\_3, 5\_4$] & 0.0 & 0.0039\\ \hline
       4 & [$0\_1, 3\_3, 5\_4$] & 0.0 & 0.004\\ \hline
       5 & [$0\_1, 3\_3, 5\_4$] & 0.0 & 0.004\\ \hline
    \end{tabular}}
  \end{center}
\end{table}

\begin{figure}[H]
\centering
\includegraphics[scale=0.5]{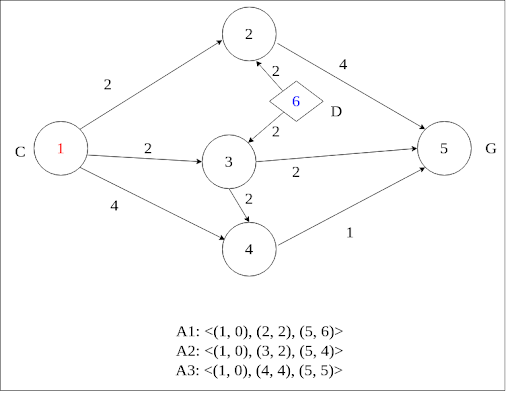}
\caption{Sample network for designing near-optimal defender strategy.}
\label{image-2}
\end{figure}

\begin{figure}[H]
\centering
\includegraphics[scale=0.36]{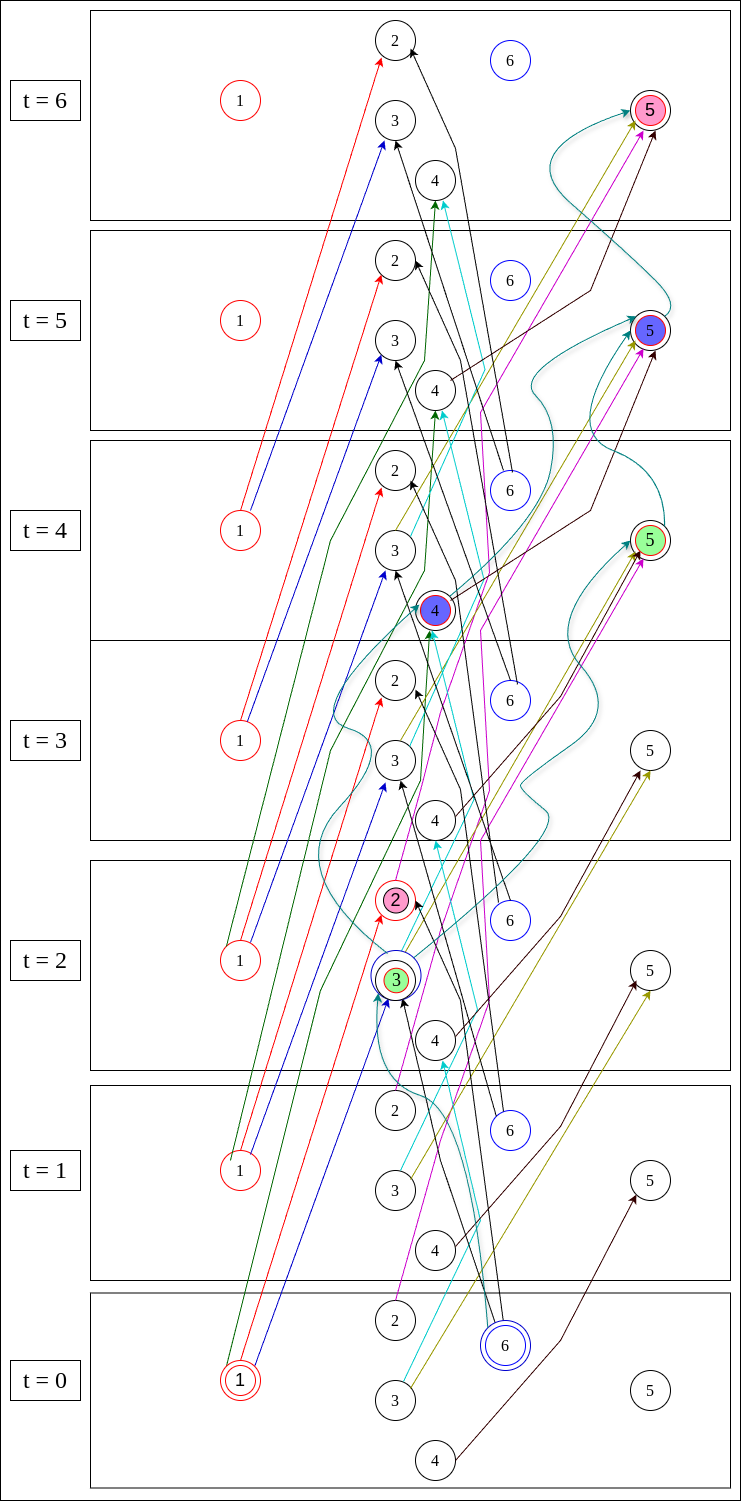}
\caption{Design of a multi-layer network for defender considering a sample network (Fig. \ref{image-2}).}
\label{image-3}
\end{figure}

In Fig. \ref{image-2}, we consider a sample network of 6 nodes in which the police station is node 6, the crime node is 1, the maximum time limit $t_{max}$ is 6, and the exit point is node 5. Here, $0\_6$ indicates node 6 at timestamp 0 $(t=0)$ in the multi-layer network. We input three attacker strategies, each with an equal probability. Each node within the same attacker strategy is colored identically in this multi-layer network. We use an approximation algorithm on the time-expanded network to generate the near-optimal defender strategy (see Fig. \ref{image-3}). The final defender strategy is represented by the green curvy lines in the multi-layer network, which follows the path $0\_6 \rightarrow 2\_3  \rightarrow 4\_4 \rightarrow 5\_5 \rightarrow 6\_5$ and $0\_6 \rightarrow 2\_3  \rightarrow 4\_5 \rightarrow 5\_5 \rightarrow 6\_5$. We demonstrate that our developed approach for the defender can generate an efficient defender strategy that interdicts all attacker strategies quickly (see Table \ref{tab:table2}).

\begin{table}[H]
  \begin{center}
   \caption{\\Defender strategy design using approximation algorithm.}
    \label{tab:table2}
    \scalebox{0.8}{
    \begin{tabular}{llll} 
    \hline
      \multicolumn{4}{c}{\textbf{Game Parameters}} \\ \hline
      \multicolumn{4}{c}{\textbf{Network Size: 6 Nodes, Crime Node: $0\_1$, Police Station: $0\_6$, $T_{max}$: 6, Exit Point: 5}}\\ \hline
      \parbox[t]{2cm}{Test Case} &  \parbox[t]{4cm}{Final Defender strategy} & \parbox[t]{4cm}{Utility of the Final\\ Defender Strategy} & \parbox[t]{2cm}{Run Time \\(Sec)}\\ \hline
       1 & [$0\_6, 2\_3, 4\_4, 5\_5, 6\_5$] & 0.0 & 0.0086\\ \hline
       2 & [$0\_6, 2\_3, 4\_5, 5\_5, 6\_5$] & 0.0 & 0.0090\\ \hline
       3 & [$0\_6, 2\_3, 4\_5, 5\_5, 6\_5$] & 0.0 & 0.0081\\ \hline
       4 & [$0\_6, 2\_3, 4\_4, 5\_5, 6\_5$] & 0.0 & 0.0083\\ \hline
       5 & [$0\_6, 2\_3, 4\_5, 5\_5, 6\_5$] & 0.0 & 0.0081\\ \hline
    \end{tabular}}
  \end{center}
\end{table}

To establish a benchmark, we compare the final defenders’ equilibrium utility generated by the developed MLN-EIGS algorithm with the final defenders’ equilibrium utility using the MILP-EIGS algorithm, which employs the exact approaches named bestDo and bestAo, developed by \cite{zhang2017optimal}. In both cases, convergence of the Stackelberg game is declared when neither player can generate a profitable deviation outside the current strategy sets, indicating that no further improving strategies exist for either player.

To assess performance on large-scale transportation networks, we consider the Central Kolkata network with 461 nodes and 1,020 edges.

\begin{table}[H]
  \begin{center}
   \caption{\\Central Kolkata Map.}
    \label{tab:table11}
    \scalebox{0.85}{
    \begin{tabular}{lccccc}
    \hline
      \textbf{Test Case} & \multicolumn{2}{c}{\textbf{Final Defender Utility}} & \textbf{Optimality Gap} &\multicolumn{2}{c}{\textbf{Run Time (Sec)}}\\
      \cline{2-3}
      \cline{5-6}
       &  MILP-EIGS &  MLN-EIGS & & MILP-EIGS &  MLN-EIGS \\ \hline
       1 & 1.0 & 1.0 & 0 &  1325.53 & 104.01\\ \hline
       2 & 1.0 & 1.0 & 0 &  1259.22 & 101.97\\ \hline
       3 & 1.0 & 1.0 & 0 &  1391.23 & 24.87\\ \hline
       4 & 1.0 & 1.0 & 0 &  1268.03 & 28.60\\ \hline
       5 & 1.0 & 1.0 & 0 &  1449.67 & 46.72\\ \hline
       6 & 1.0 & 0.0 & 1 &  1255.55 & 141.99\\ \hline

    \end{tabular}}
  \end{center}
\end{table}

\begin{figure}[H]
\centering
\includegraphics[scale=0.5]{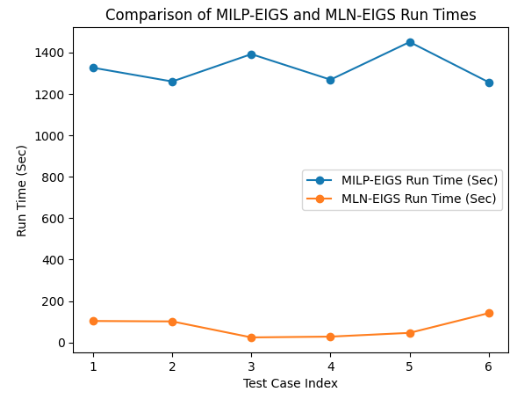}
\caption{Run Time Analysis: MILP-EIGS vs. MLN-EIGS.}
\label{image-5}
\end{figure} 

Table \ref{tab:table11} reports the computational performance of MLN-EIGS and MILP-EIGS across several test instances on the Central Kolkata network. The performance gap becomes significant for the Central Kolkata network, where MLN-EIGS completes within approximately 100 seconds, compared to nearly 20 minutes for MILP-EIGS (see Fig. \ref{image-5}). This substantial disparity highlights the scalability limitations of the MILP-based approach. Since interdiction problems are inherently time-sensitive—particularly when the objective is to interdict an attacker within a limited time horizon—computational efficiency is critical. Although MILP-EIGS is capable of producing optimal solutions, its excessive runtime renders it impractical for large-scale transportation networks. Notably, for the considered network, most test cases exhibit a zero optimality gap between MLN-EIGS and MILP-EIGS, indicating that MLN-EIGS achieves solutions of equivalent quality. However, a substantial difference is observed in computational time. Overall, these results demonstrate that MLN-EIGS consistently delivers high-quality solutions with significantly improved computational efficiency, while MILP-EIGS struggles to provide time-efficient performance for the proposed Stackelberg game formulation.

\section{Conclusion}
\label{S:7}

This paper proposes a time-dependent Stackelberg interdiction framework on a time-expanded network that captures dynamic movement, temporal feasibility constraints, and probabilistic interception within a unified modeling structure. In contrast to classical static security games and deterministic network interdiction models, the proposed approach explicitly incorporates temporal layering and sequential interception risk through a multiplicative escape-probability formulation. To address the resulting nonlinear structure of the follower’s optimization problem, a logarithmic transformation is employed, enabling efficient computation of the attacker’s best response via shortest-path techniques. This reformulation significantly enhances computational scalability compared to conventional MILP-based approaches.
The proposed framework therefore establishes a systematic connection between dynamic network modeling and Stackelberg security games, offering both probabilistic consistency in interception modeling and computational tractability for large-scale, time-dependent transportation networks. Computational experiments on real transportation networks demonstrate that the proposed method substantially reduces computational time while achieving the same defender utility as the benchmark MILP formulation.
Despite these contributions, certain limitations remain. In particular, the current model does not incorporate real-time traffic dynamics or stochastic travel conditions. Future research may focus on integrating traffic flow variability into the time-expanded framework and developing exact strategy-generation procedures for the defender within the layered network structure.
\\
\textbf{Acknowledgments}
We are grateful to the members of the Multi-Agent Laboratory at Kyushu University for their insightful discussions and comments. This research is funded by a project supported by the Grants-in-Aid for Scientific Research from the Japan Society for the Promotion of Science.

{}
\end{document}